\documentclass{article}

    \usepackage[preprint]{neurips_2020}

\usepackage[utf8]{inputenc} 
\usepackage[T1]{fontenc}    
\usepackage{hyperref}       
\usepackage{url}            
\usepackage{booktabs}       
\usepackage{amsfonts}       
\usepackage{nicefrac}       
\usepackage{microtype}      
\usepackage{xcolor}         
\usepackage{graphicx}       
\usepackage{subfigure}      
\usepackage{amsmath}        
\usepackage[title]{appendix}

\title{NP-PROV: Neural Processes with Position-Relevant-Only Variances}

\author{
  Xuesong Wang \\
  University of New South Wales \\
  \texttt{xuesong.wang1@student.unsw.edu.au} \\
   \And
   Lina Yao \\
   University of New South Wales \\
   \texttt{lina.yao@unsw.edu.au} \\
   \AND
   Xianzhi Wang \\
   University of Technology Sydney \\
   \texttt{xianzhi.wang@uts.edu.au} \\
   \And
   Feiping Nie \\
   Northwestern Polytechnical University \\
   \texttt{feipingnie@gmail.com} \\
}

\begin{document}

\maketitle

\begin{abstract}
Neural Processes (NPs) families encode distributions over functions to a latent representation, given context data, and decode posterior mean and variance at unknown locations. Since mean and variance are derived from the same latent space, they may fail on out-of-domain tasks where fluctuations in function values amplify the model uncertainty. We present a new member named Neural Processes with Position-Relevant-Only Variances (NP-PROV). NP-PROV hypothesizes that a target point close to a context point has small uncertainty, regardless of the function value at that position. The resulting approach derives mean and variance from a function-value-related space and a position-related-only latent space separately. Our evaluation on synthetic and real-world datasets reveals that NP-PROV can achieve state-of-the-art likelihood while retaining a bounded variance when drifts exist in the function value. 
\end{abstract}

\section{Introduction}
Neural networks (NNs) are proven effective in various machine learning tasks for making deterministic predictions. They have the flexibility of describing and fitting any type of data distributions. Nevertheless, most neural networks are incapable of evaluating model stochasticity. They cannot handle tasks where prediction uncertainty is equally crucial, e.g., autonomous driving \cite{5979587auto-driving}, disease diagnosis \cite{lorenzi2019probabilistic-disease}, and stock market forecasting \cite{christou2017economic-stock}. In contrast, Gaussian processes (GPs) model data with an infinite sequence of correlated normal distributions and are intrinsically suitable for such problems. Conditioned on some prior knowledge, e.g., context points with positions \textbf{X} and function values \textbf{Y}, they are able to infer the likelihood of target values \(\textbf{Y}_\star\) at unknown locations \(\textbf{X}_\star\). Despite the advantages, GPs require designing data-specific kernel functions; the cubic computational cost of matrix inversion impedes GPs from handling large-scale data.

The recent progress in models based on Neural Process (NP) \cite{garnelo2018neural} have brought GPs the advantages of fast forward propagation and powerful feature representations of NNs. Basic NPs represent a stochastic process with an encoder-decoder network structure. It encodes context data \((\textbf{X ,Y})\) to a latent representation \(\textbf{Z} \sim \textbf{P}(\textbf{Z}| \textbf{X}, \textbf{Y})\) and decodes it to a posterior probability of the target data based on its relationship with the context points \(\textbf{Y} \sim \textbf{P}(\textbf{Y}_\star | \textbf{Z};\textbf{X}_\star)\). Recent years have witnessed a family of NP-variants, such as Attentive NP \cite{kim2019attentive}, Convolutional Conditional NP \cite{gordon2019convolutional}, and Sequential NP \cite{singh2019sequential}. They improve NP via aggregating context knowledge non-linearly and considering spatial or temporal relationships among target points but still decode mean and variance from the same latent variable---meaning that their variances are correlated to \textbf{Y}. When future testing sets have shifts compared to training sets (e.g., stock market price with incremental trends), fluctuations in \textbf{Y} can severely amplify model uncertainty. A model can be stable on out-of-domain tasks only when the variance inference is relevant to locations \(\textbf{X}\) yet irrelevant to function values \(\textbf{Y}\).

In this paper, we introduce a new member named Neural Processes with Position-Relevant-Only Variances (NP-PROV), which derives mean and variance functions from two coupled latent spaces. Mean values are related to function values \textbf{Y}, self-correlation within context locations \(\textbf{X}\), and cross-correlations between context positions \textbf{X} and target positions \(\textbf{X}_\star\). Variance values exclude function values yet are also associated with the self-correlations within the target positions. Our main contributions are as follows:

\begin{itemize}
    \item We designed an auto-encoder module associated with self-correlations between data points. It reconstructs context data through the module and reduces model uncertainty in scenarios where context points have higher self-correlations.
    \item The approach derives mean and variance values from two coupled latent spaces. The position-relevant-only variances prevent the model from estimating oscillating uncertainty when function values are out of the training range.
    \item The proposed model achieves state-of-the-art performance over on-the-grid and off-the-grid datasets. Specifically, three GP-based kernels and a real-world time series are adopted to sample off-the-grid tasks. Four image inpainting tasks are evaluated: MNIST, SVHN, celebA and miniImageNet.
\end{itemize}

\section{Background}
\label{background}
\paragraph{Gaussian Processes.} Let  \(\textbf{Y} = [\textnormal{y}_{1}, ..., \textnormal{y}_{N}]^\top \in \mathbb{R}^{N}\) be a set of \emph{N} observations   at input locations \(\textbf{X} = [\textnormal{x}_{1}, ..., \textnormal{x}_{N}]^\top \in \mathbb{R}^{N \times D}\), a function \(f: \mathbb{R}^D \mapsto \mathbb{R}\), and a likelihood \(p(\textbf{Y| F};\textbf{X})\), where \(\textbf{F} = f (\textbf{X}) \) denotes the function values at the input locations. A Gaussian process (GP) prior is placed on the function \(f\); it models all function values as a jointly Gaussian distribution to infer \(f\). The prior has a mean function \( m(\cdot): \mathbb{R}^D \mapsto \mathbb{R} \) and a covariance function \(\mathcal{K}(\cdot , \cdot): \mathbb{R \times R} \mapsto \mathbb{R} \). The generative model of the corresponding GP is as follows:

\begin{equation}
 p(\textbf{Y| F};\textbf{X}) = \mathcal{N}( m(\textbf{X}),  \mathcal{K}( \textbf{X}, \textbf{X}^\top) )    
\end{equation}

GP hypothezes similar function values of two inputs that are highly correlated. Given some context samples (\textbf{X}, \textbf{Y}), we can perform probabilistic inference and assign posterior distributions over unknown locations from the same function. For new input locations \(\textbf{X}_\star=[\textnormal{x}_{\star 1}, ..., \textnormal{x}_{\star M}]^\top \),  the posterior distribution also follows a joint Gaussian distribution \( p(\textbf{Y}_\star | \textbf{F};\textbf{X}_\star, \textbf{X}, \textbf{Y}) = \mathcal{N}( \mu_\star, \Sigma_\star )\):

\begin{equation}
    \label{equ: GP}
    \begin{gathered}
    \mu_\star = m(\textbf{X}_\star)+ \textbf{K}_\star^\top \textbf{K}^{-1}(\textbf{Y} - m(\textbf{X})) \\
    \Sigma_\star = \textbf{K}_{\star \star} - \textbf{K}_\star^\top \textbf{K}^{-1} \textbf{K}_\star
    \end{gathered}
\end{equation}
where \(\textbf{K}= \mathcal{K}( \textbf{X}, \textbf{X}^\top), \textbf{K}_\star = \mathcal{K}( \textbf{X}, \textbf{X}_\star^\top)\), and \(\textbf{K}_{\star \star} =  \mathcal{K}( \textbf{X}_\star, \textbf{X}_\star^\top) \). Intuitively, when there is no trend in observations (i.e., \(m(\textbf{X}_\star) =m(\textbf{X}) = 0\)), the mean of \( \textbf{Y}_\star \) is a linear combination of elements in \( \textbf{Y} \). The weights of those elements are associated with the self-correlation of \(\textbf{X}\) and cross-correlation between \( \textbf{X}\) and \( \textbf{X}_\star\). The variance equals the total uncertainty of \(\textbf{X}_\star\) minus the certainty induced from \(\textbf{X}\) and is irrelevant to function values \( \textbf{Y}\).

\paragraph{Convolutional Conditional Neural Processes.} 
As a type of latent variable model, NPs inherit distribution consistency from GPs, where context data (\textbf{X, Y}) and target data (\(\textbf{X}_\star, \textbf{Y}_\star\)) are sampled from the same function and share latent variables.

\begin{figure}[ht]
    \centering
        \subfigure[]
        {\includegraphics[width=0.35\linewidth]{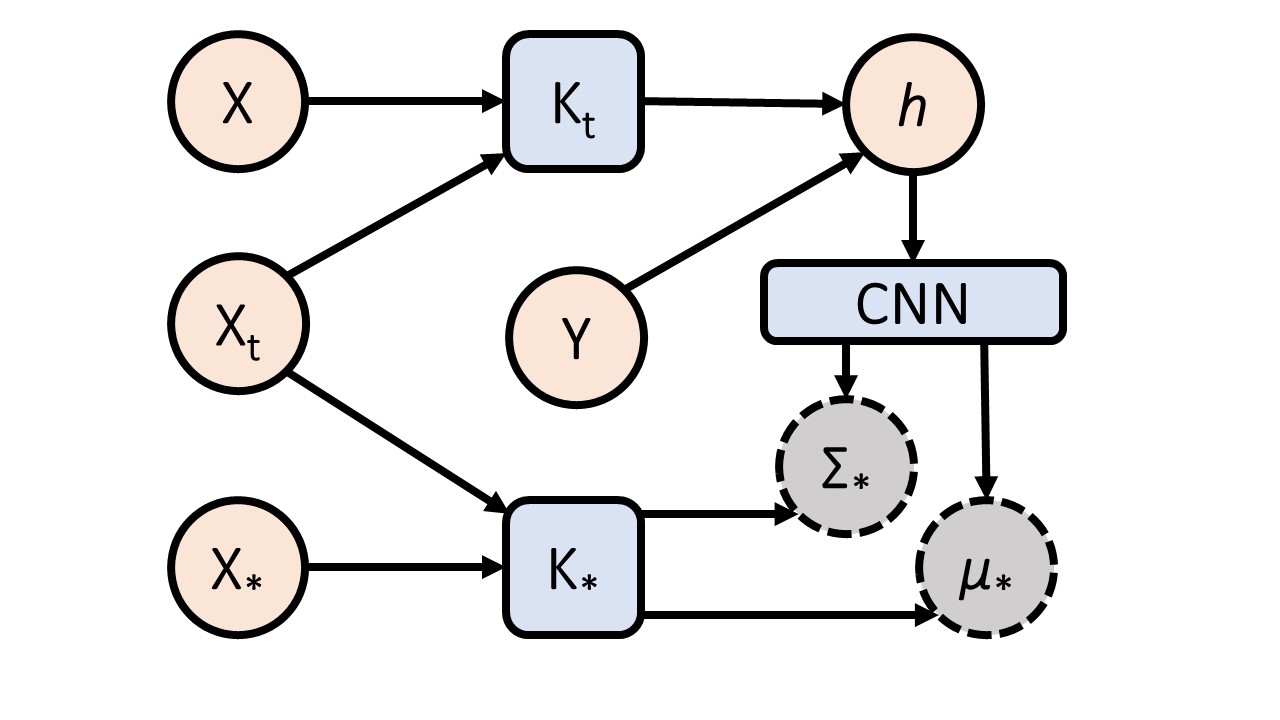}}
        \subfigure[]
        {\includegraphics[width=0.25\linewidth]{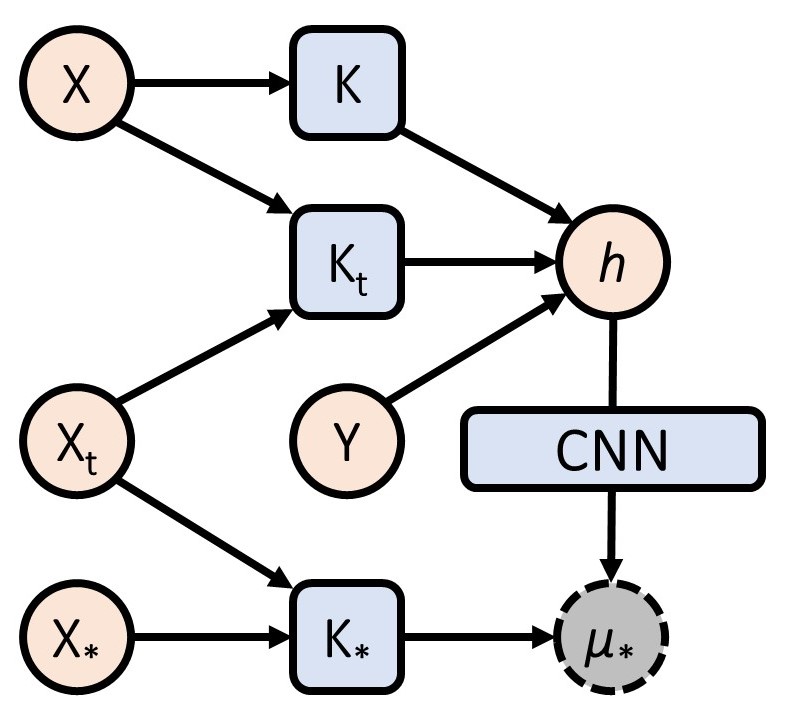}}
        \subfigure[]        {\includegraphics[width=0.25\linewidth]{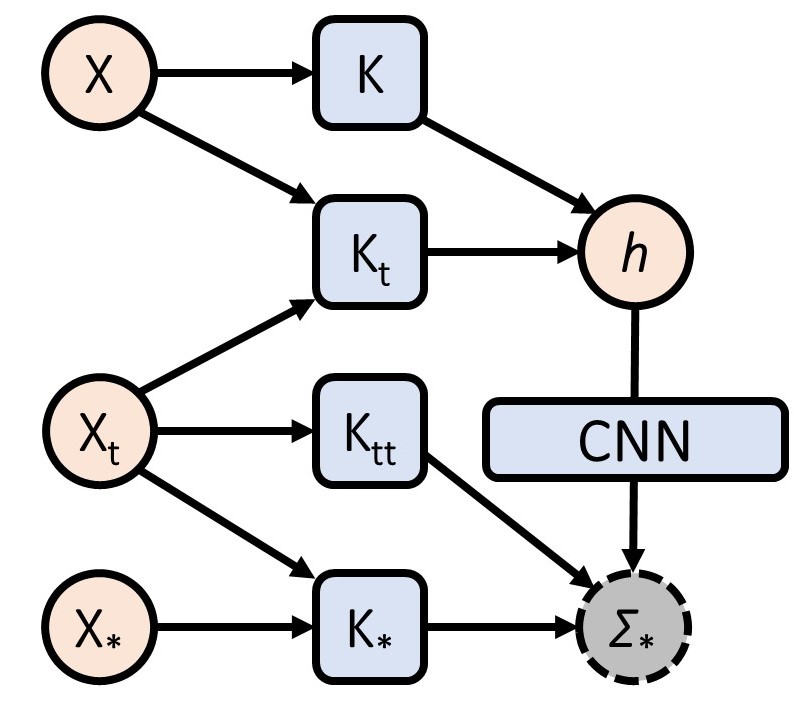}}
    \caption{Graphical Models for (a) ConvCNP (b) NP-PROV mean and (c) NP-PROV variance}
    \label{fig:graphical model}
\end{figure}

Convoluational Conditional Neural Processes (ConvCNP) differ from other NP families in handling testing data outside the range of training data. They introduce a discretization of a continuous range of \(\textbf{X} \cup \textbf{X}_\star\) named \(\textbf{X}_t\). Convolution operations enable NPs to focus on local receptive fields of gridded inputs and to generalize to out-of-domain predictions.
The posterior distribution of \( p(\textbf{Y}_\star | \textbf{X}_\star, \textbf{X}_t, \textbf{X}, \textbf{Y}) = \mathcal{N}( \mu_\star, \Sigma_\star )\) is also in the form of a Gaussian distribution:

\begin{equation}
    \begin{gathered}
     \mu_\star = \psi_{\star\mu}\{\mathbf{K}_{\star}^\top \mathbf{CNN}[ \psi_t(\mathbf{K}_t^\top \mathbf{Y})]\} \\
     \Sigma_\star = \psi_{\star\Sigma}\{\mathbf{K}_{\star}^\top \mathbf{CNN}[ \psi_t(\mathbf{K}_t^\top \mathbf{Y})]\}
    \end{gathered}
\end{equation} 
where \( \mathbf{K}_\star = \mathcal{K}_\star(\mathbf{X}_\star, \mathbf{X}_t^\top), \mathbf{K}_t = \mathcal{K}_t(\mathbf{X}_t, \mathbf{X}^\top) \) are covariance functions with learnable length scales; \(\psi_t, \psi_{\star\mu}\) and \(\psi_{\star\Sigma}\) are positive-definite transformations associated with a Reproducing Kernel Hilbert Space; the graphical model of ConvCNP is described in Fig \ref{fig:graphical model} (a); \(\psi_t(\mathbf{K}_t^\top \mathbf{Y})\) encodes prior knowledge about the function distribution on the entire grid \(\mathbf{X}_t\); \(\mathbf{CNN}(\cdot)\) enables model to exhibit translation invariance; \( \psi_{\star\mu}(\mathbf{K}_{\star}^\top\cdot)\) and \( \psi_{\star\Sigma}(\mathbf{K}_{\star}^\top\cdot)\) translate the knowledge to predict target mean and standard deviation for \(p(\mathbf{Y}_\star)\).

\section{Neural Processes with Position-Relevant-Only Variances}
\label{Methodology}
\(\textbf{K}_t\) evaluates correlation between \(\textbf{X}_t\) and \textbf{X}, and \(\textbf{K}_\star\) evaluates cross correlation between \(\textbf{X}_\star\) and \(\textbf{X}_t\). Therefore, \(\psi_t(\textbf{K}_t\textbf{Y})\) and  \( \psi_{\star}(\mathbf{K}_{\star}^\top\cdot)\) in ConvCNP act as two cascaded cross-correlation modules in GPs. We propose a new member Neural Processes with Position-Relevant-Only Variances (NP-PROV) to improve ConvCNP in two ways:

1. Considering self-correlations within context data. ConvCNP usually considers cross-correlations between target data and context data only. It is possible to further reduce the model uncertainty near the region where context data have high self-correlations.

2. Changing variance inference to position-relevant-only. The variance derivation is related to function values in ConvCNP. It can effectively prevent the model from amplifying uncertainty caused by value fluctuations if it is made to be only associated with relative distances among locations (as in GP).

In the following, we detail in an off-the-grid scenario with continuous inputs and an on-the-grid scenario with intrinsically discretized inputs.

\subsection{Off-the-grid Scenario.}

\paragraph{Mean function}
We follow convCNP to construct the model with two cascaded GP-alike layers. The first layer maps \textbf{X} to a uniformly discretized grid space \(\textbf{X}_t = [\textnormal{x}_1, ..., \textnormal{x}_t]^\top\) built on the lower and upper range of \(\textbf{X} \cup \textbf{X}_\star\). The second layer maps the space back to \(\textbf{X}_\star\). As illustrated in Fig \ref{fig:graphical model}(b), an auto-encoder structure is computed to match \(\textbf{K}^{-1}\textbf{Y}\) first:

\begin{equation}
    \begin{gathered}
        h^{self}_{ni} = \psi_E(\textnormal{K}_{ni} y_n)\\
        \Tilde{y}_i = \psi_D(\sum_{n=1}^N h^{self}_{ni}\textnormal{K}_{ni}), \quad
        \mathcal{L}_2 = \frac{1}{N}\sum_{i=1}^N(y_i - \Tilde{y}_i)^2
    \end{gathered}
\end{equation}
where \(h^{self}_{ni}\) is self-correlation mapping from the \(n\)-th context point to the \(i\)-th context point; \(\textnormal{K}_{ni}\) represents corresponding self-covariance elements in \(\textbf{K} = \mathcal{K} (\textbf{X}, \textbf{X}^\top) \in \mathbb{R}^{n \times n}\); \(\psi_E\) and \(\psi_D\) are positive-definite linear transformations. Since \(\textbf{K}\textbf{K}^{-1}\textbf{Y} = \textbf{Y}\), we use auto-encoder structure to map \(h^{self}_{ni}\) back to \(y_i\) and minimize reconstruction loss \(\mathcal{L}_2\) to mimic matrix inversion. Then, we can compute cross-correlation weights associated with \(\textbf{K}^t = \mathcal{K} (\textbf{X}_t, \textbf{X}^\top) \in \mathbb{R}^{t\times n}\):
\begin{equation}
    \begin{gathered}
        h^{cross}_j = \sum_{n=1}^N[ 1 ,\quad y_n ]^T\textnormal{K}^{t}_{nj}, \quad h_j^{cross (1)} = h_j^{cross (1)}/ h_j^{cross (0)}\\
        h_j = \psi_t([ \frac{1}{N^2} \sum_{i=1}^{N}\sum_{n=1}^{N}h^{self}_{ni}, \quad h^{cross}_j ]) \\
    \end{gathered}
\end{equation}
where \( h^{cross}_j \) maps the \(n\)-th context point to the \(j\)-th data in \(\textbf{X}_t\). \(\textbf{I}  = \textbf{Y}^0\) is an additional identity density channel appended to \(\textbf{Y} \leftarrow [\textbf{I} \; \textbf{Y}]\) and is the powers of Y up to order-1 for normalizing large variance. Dividing \(h_i^{(0)}\) helps to normalize scales when there are large variations in \textbf{X}. A new linear transformation \(\psi_t\) fuses averaged self-correlation and cross-correlation information to a latent variable \(h_j\) on the \(j\)-th data of \(\textbf{X}_t\).

A CNN takes the obtained condition \(h_j\) to suffice translation invariance.It uses a UNet struture with skipped concatenation of 6 convolution layers and 6 transposed convolution layers to capture global and local patterns. The CNN enables NP-PROV to predict out-of-domain tasks by handling locations outside the training range through filter sliding. When projecting the output back from \(\textbf{X}_t\) to \(\textbf{X}_\star\), \(\psi_{\star}\) transforms the cross-correlation weights associated with \(\mathbf{K}^\star = \mathcal{K}(\textbf{X}_\star, \textbf{X}_t^\top) \in \mathbb{R}^{m\times t}\). The overall mean function of the \(m\)-th target point in \(\textbf{X}_\star\) is as follows:
\begin{equation}
\label{CNN for 1d}
  \mu^\star_{m} = \psi^\star_{\mu}(\sum_{j=1}^t \textbf{CNN}(h_j) \textbf{K}^\star_{jm})
\end{equation}

\paragraph{Covariance function}
Based on Eq \ref{equ: GP}, we replace original values \textbf{Y} with \(\textbf{K}^t\) in covariance functions and add a module associated with self-correlations within the grided data \(\textbf{X}_{t} \). As in Fig \ref{fig:graphical model} (c), we define two new mappings: \(\psi\) and \(\psi_{tt}\):
\begin{equation}
    \begin{gathered}
    h_i = \psi(\sum_{j=1}^{t}\textbf{K}^{t^\top}_{ji}), \qquad
    h^{self}_j = \psi_{tt}(\sum_{k=1}^{t}\textbf{K}^{tt}_{kj})
    \end{gathered}
\end{equation}
where \(\mathbf{K}^{t^\top} = \mathcal{K} (\textbf{X}, \textbf{X}_t^\top) \in \mathbb{R}^{n\times t}\); \(\textbf{K}^{tt} = \mathcal{K} (\textbf{X}_t, \textbf{X}_t^\top) \in \mathbb{R}^{t \times t}\). \(h_i\)  maps \(\textbf{X}_t\) to \textbf{X} and replaces values \textbf{Y} in the mean function. This value-irrelevant input will generate a new \(h_j\). After the convolution on the new \(h_j\), the result is concatenated with \(h_{j}^{self}\). The overall covariance function is as follows:
\begin{equation}
  \Sigma^\star_m = \psi^\star_{\Sigma}(\sum_{j=1}^t[ \textbf{CNN}(h_j), \quad h^{self}_{j} ] K^\star_{jm})
\end{equation}

Again, as in Fig\ref{fig:graphical model}(c), the added elements in NP-PROV can retain reasonable variance regardless of drifts in function values when compared with ConvCNP.

\paragraph{Inference and Learning}
With \(\mu_\star\) and \(\Sigma_\star\), the posterior distribution of \(\textbf{Y}_\star\) follows a multivariate normal distribution: \(\mathcal{N}(\mu_\star, \Sigma_\star)\). The training objective is to maximize the log-likelihood of target outputs meanwhile minimizing the reconstruction loss, given all parameters defined as \(\Theta\):
\begin{equation}
  \theta^\star = \arg \max_{\theta\in\Theta} \sum_{m=1}^M \log p(y_m |\mathcal{N}(\mu_\star,  \Sigma_\star)) - \frac{1}{N}\sum_{i=1}^N(y_i - \Tilde{y}_i)^2
\end{equation}

\subsection{On-the-grid Scenario.}
CNN can be applied to On-the-grid data (e.g., images), which are discretized and represent better spatial correlations between context and target data with little efforts. We describe the context inputs \(\textbf{X}\) using a context mask \(\textbf{M}\), where an element value is 1 if a pixel is revealed and 0 otherwise. The target inputs \(\textbf{X}_\star\) formulate the whole image.

\paragraph{Mean function}
Similar to off-the grid data, self-correlations and cross-correlations weights are computed for the latent variable extraction. Instead of adopting kernel functions, convolutions are implemented on the context masks \(\textbf{M}\) and the context values \(\textbf{M}  \odot \textbf{Y}\):
\begin{equation}
    \begin{gathered}
     h_{self} = \psi_{E}(\textbf{M}), \quad \Tilde{M} = \psi_{D}(h_{self}),  \qquad  \mathcal{L}_2 =  \parallel \textbf{M} - \Tilde{\textbf{M}} \parallel_{2}\\
    h_{cross} = \psi([\textbf{M}, \quad  \textbf{M}  \odot \textbf{Y}]) \\
    \end{gathered}
\end{equation}
where \textbf{Y} represents a full image, \(\psi_E, \psi_D, \psi\) are 1-layer convolution operations. \(\psi_E\) aims to extract self-correlations within unmasked points. \(\psi\) shows cross correlations within unmasked pixel values. Then, a UNet CNN structure maps the concatenated latent variable to a translation invariant representation. Finally, a new convolution \(\psi_{\star\mu}\) maps the representation to the posterior mean: 

\begin{equation}
  \mu_\star = \psi_{\star\mu}(\textbf{CNN}( [h_{self} \quad h_{cross}] ))
\end{equation}

\paragraph{Covariance function}
We substitute the masked image \(\textbf{M}  \odot \textbf{Y}\) with the mask \(\textbf{M}\) to extract a output-irrelevant latent variable \(h\) for variance functions:
\begin{equation}
    \begin{gathered}
     h_{cross} = \psi([\textbf{M}, \quad  \textbf{M}])  \quad  h = [h_{self} \quad h_{cross}]\\
    \end{gathered}
\end{equation}

Also, we add a self-correlation layer on the target masks, i.e., an identity matrix \textbf{I} :
\begin{equation}
    h_{\star\star} = \psi_{\star\star}(\textbf{I})
\end{equation}

The overall function learns a new transformation \(\psi_{\star\Sigma}\) that maps \(h\) and \(h_{\star\star}\) to to the posterior covariance. The objective is to maximize a log-likelihood to recover the whole image and to minimize the reconstruction error of the mask simultaneously.

\begin{equation}
  \Sigma_\star = \psi_{\star\Sigma}([\textbf{CNN}(h) \quad h_{\star\star}])
\end{equation}

\section{Experiments}
\label{Experiments}
 We conduct few-shot regressions tasks on off-the-grid 1d datasets and on-the-grid images to evaluate the effectiveness of NP-PROV.
 
\subsection{Off-the-grid datasets}
We aim to maximize the likelihood of the outputs \(\textbf{Y}_\star\) at unknown locations \(\textbf{X}_\star\), given observations \(\textbf{Y}\) at input locations \(\textbf{X}\). We are interested in the following questions: (a) Are the prediction and uncertainty estimation reasonable? (b) How will the model perform when the testing range of X and Y goes beyond the training data, i.e., out-of-domain tasks? (c) How will the self-correlation affect model prediction? We compare the results of four state-of-the-art neural process members: Neural Process (NP) \cite{garnelo2018neural}, Conditional Neural Process (CNP) \cite{garnelo2018conditional}, Attentive Neural Process (ANP) \cite{kim2019attentive}, and Convolutional Conditional Neural Process(ConvCNP)\cite{gordon2019convolutional}.
We use three challenging kernels to generate synthesised Gaussian processes functions, according to \cite{gordon2019convolutional}:
\begin{itemize}
    \item \(\textnormal{EQ}: \quad \mathcal{K}(x, x^\prime) = e^{-\frac{1}{2}(\frac{x - x^\prime}{0.25})^2}\)
    \item Matern \(-\frac{5}{2}\): \quad \(\mathcal{K}(x, x^\prime)=(1+4\sqrt{5}d + \frac{5}{3}d^2)e^{-\sqrt{5}d}\) with \(d = 4|x - x^\prime|\)
    \item Weakly periodic: \( \mathcal{K}(x, x^\prime) = e^{-\frac{1}{2}(f_1(x) - f_1(x^\prime))^2 -\frac{1}{2}(f_2(x) - f_2(x^\prime))^2} \cdot e^{-\frac{1}{8}(x - x^\prime)^2}\), with \(f_1(x)= \cos({8\pi x}\)) and \(f_2(x)= \sin({8\pi x})\)
\end{itemize}
The training data range within \(x\in\) [-2, 2] for GP-sampled datasets. Also, a real-world time series dataset Smart Meter \cite{3springs2019github} is added. It contains energy consumption readings from a sample of 5,567 London Households between November 2011 and February 2014. We select timestamp in days as the input \textbf{X} and consumption in kWh/ half-hour as the output \textbf{Y}. The x range is set to [0, 2], representing 
a relative 2-day window from a random clip on the time axis. The number of context points and target points in each task are uniformly distributed in \(\mathcal{U}(3, 50)\). A batch of 16 tasks is generated and the total number of tasks in one epoch is 256. We train each model for 200 epochs and then test them using 2,048 new generated tasks for 6 times. 

\begin{figure}[ht]
    \centering
        {\includegraphics[width=0.32\linewidth]{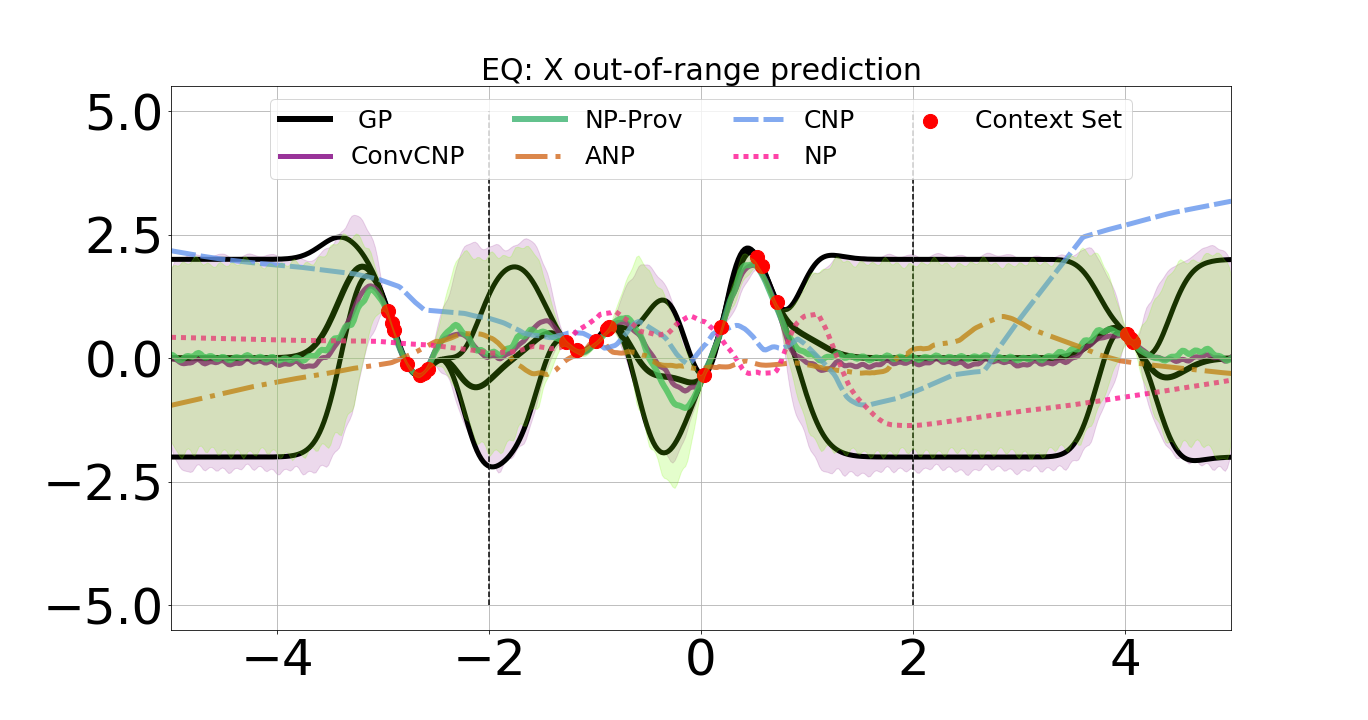}}
        {\includegraphics[width=0.32\linewidth]{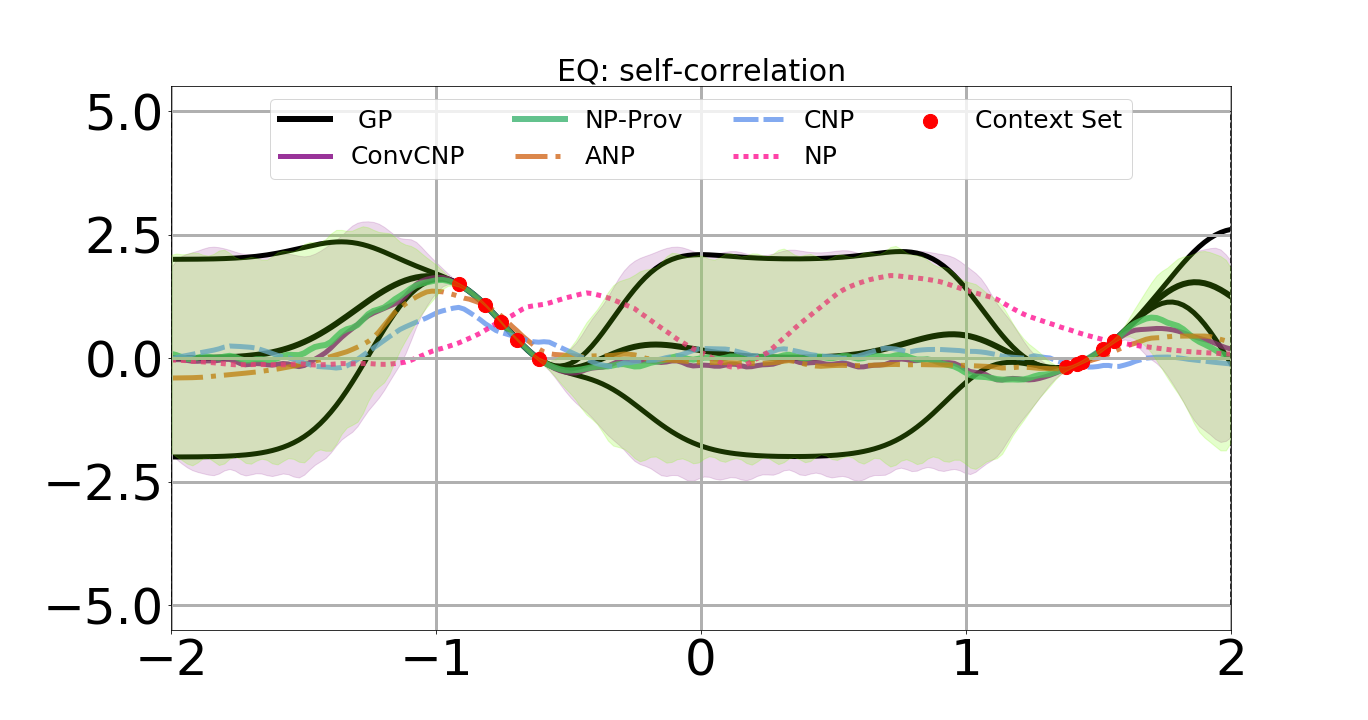}}
        {\includegraphics[width=0.32\linewidth]{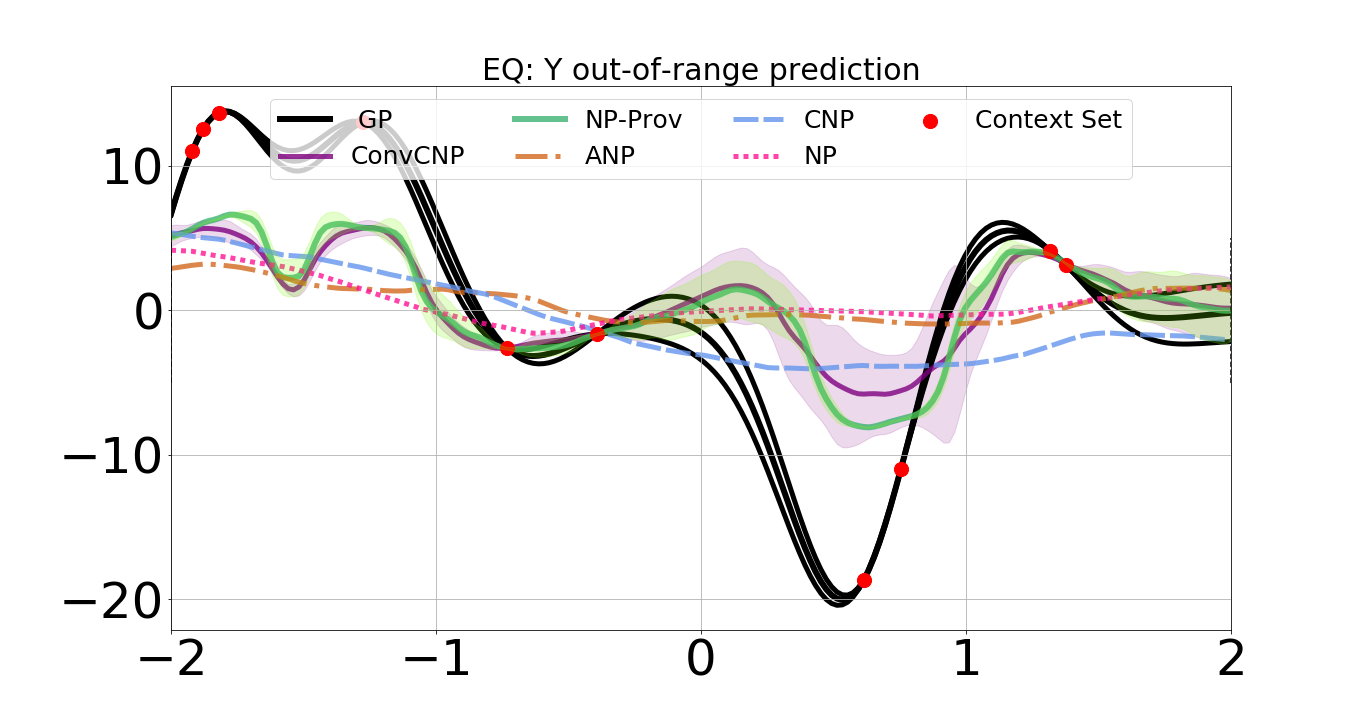}}
        {\includegraphics[width=0.32\linewidth]{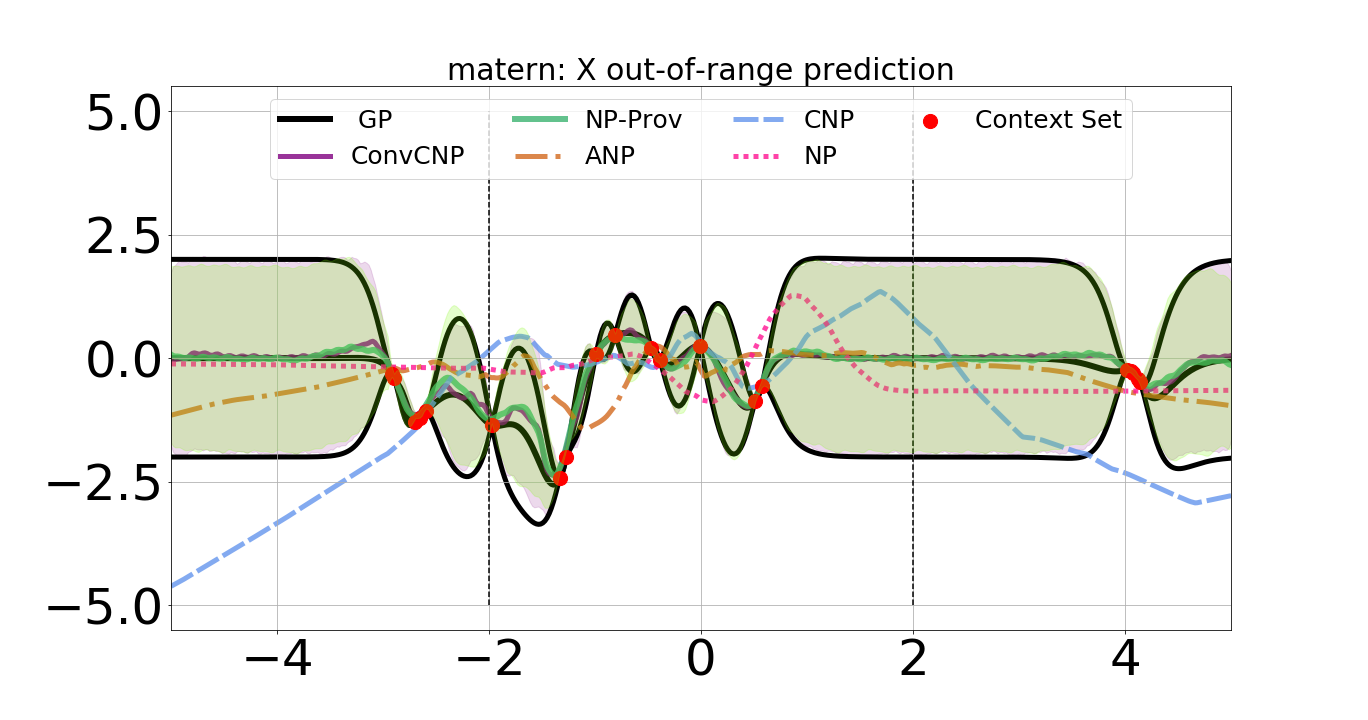}}
        {\includegraphics[width=0.32\linewidth]{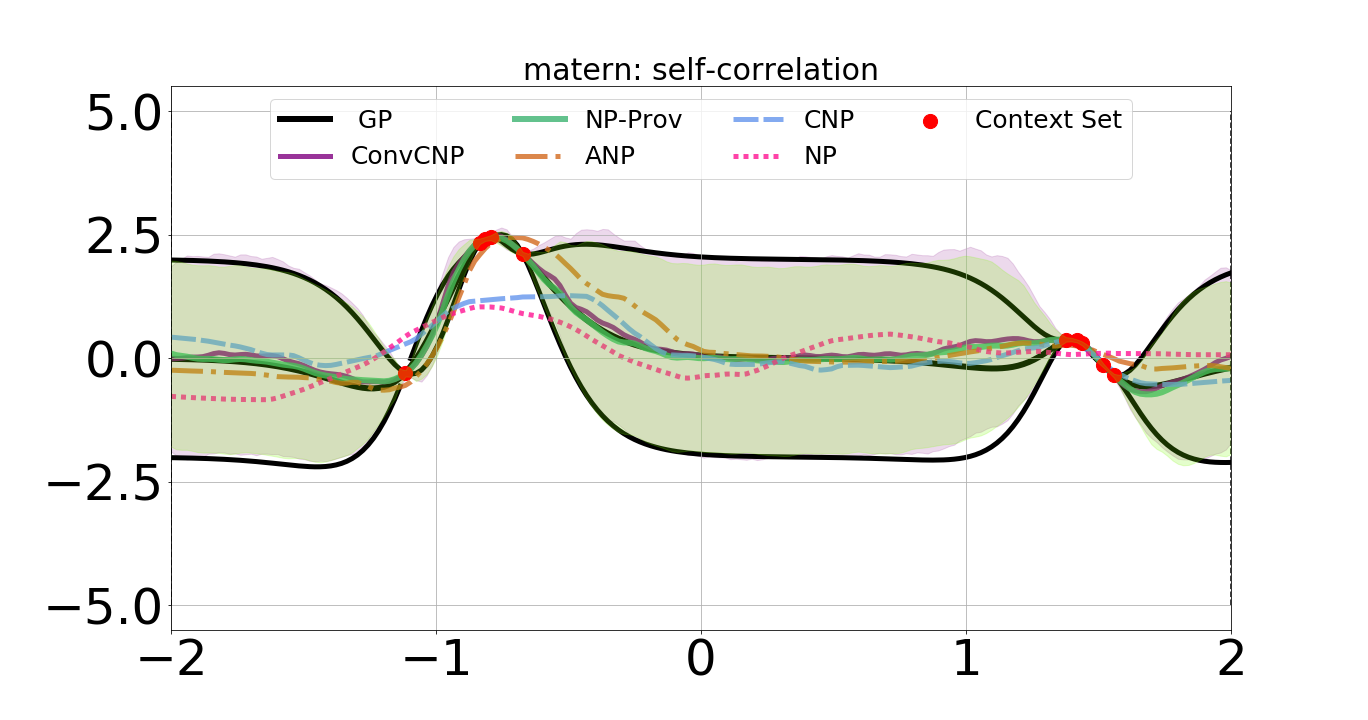}}
        {\includegraphics[width=0.32\linewidth]{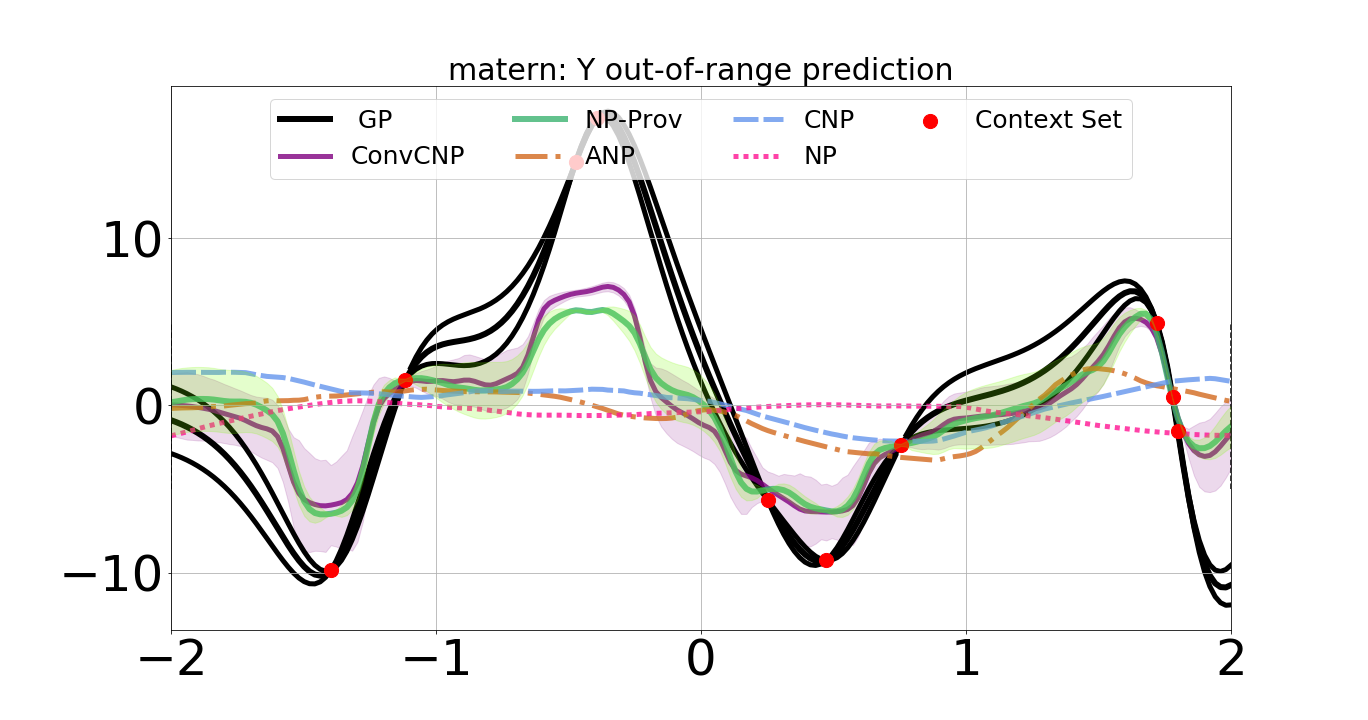}}
        {\includegraphics[width=0.32\linewidth]{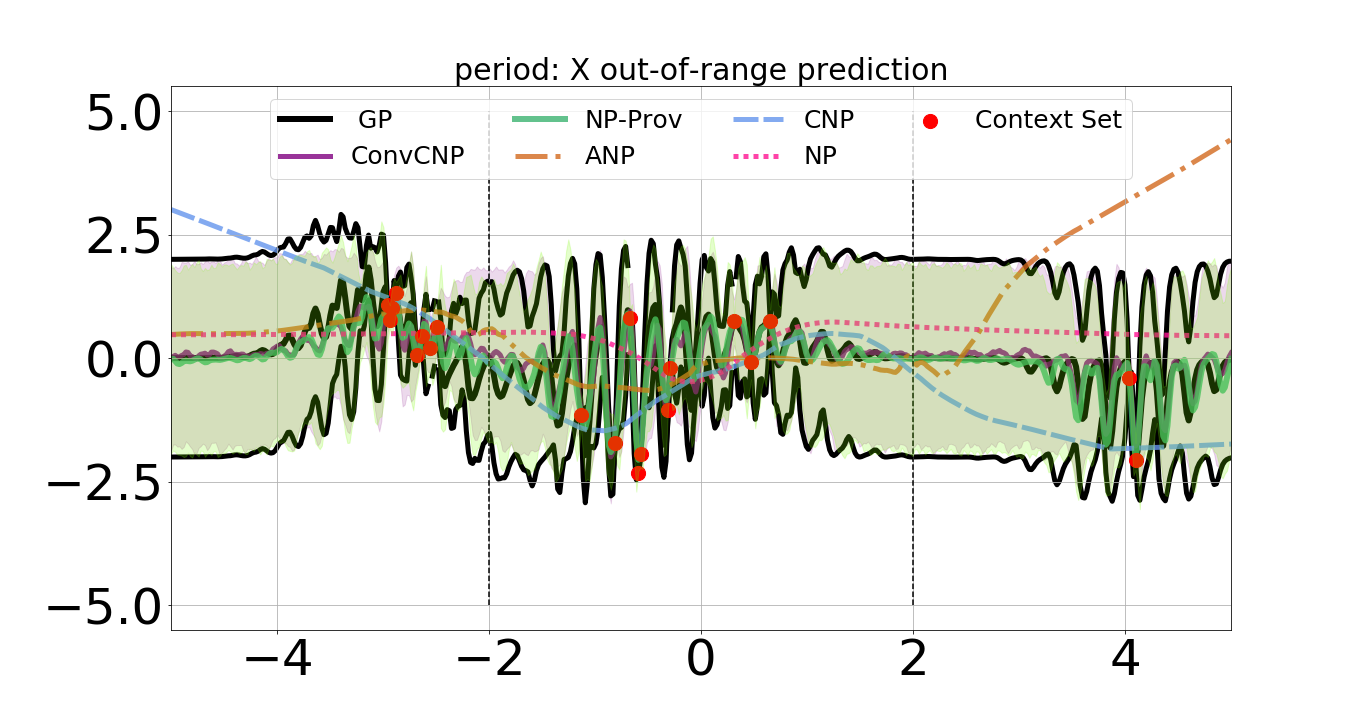}}
        {\includegraphics[width=0.32\linewidth]{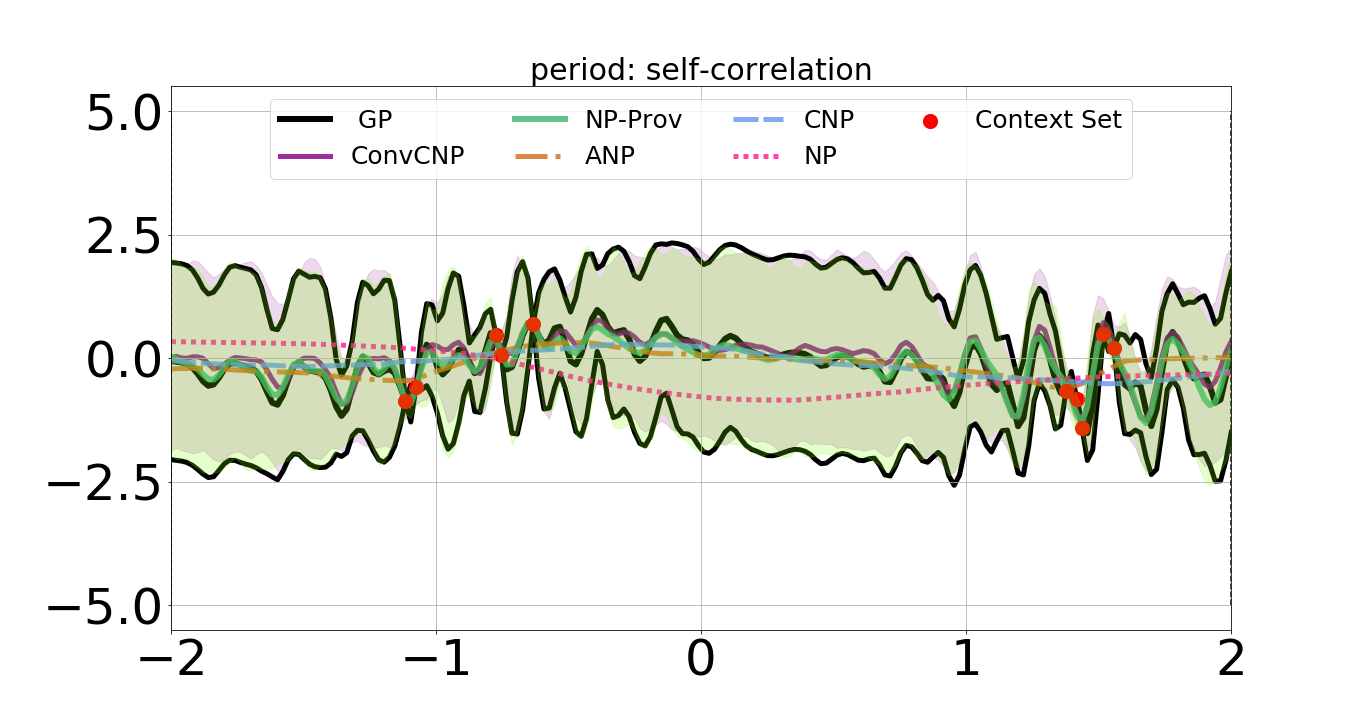}}
        {\includegraphics[width=0.32\linewidth]{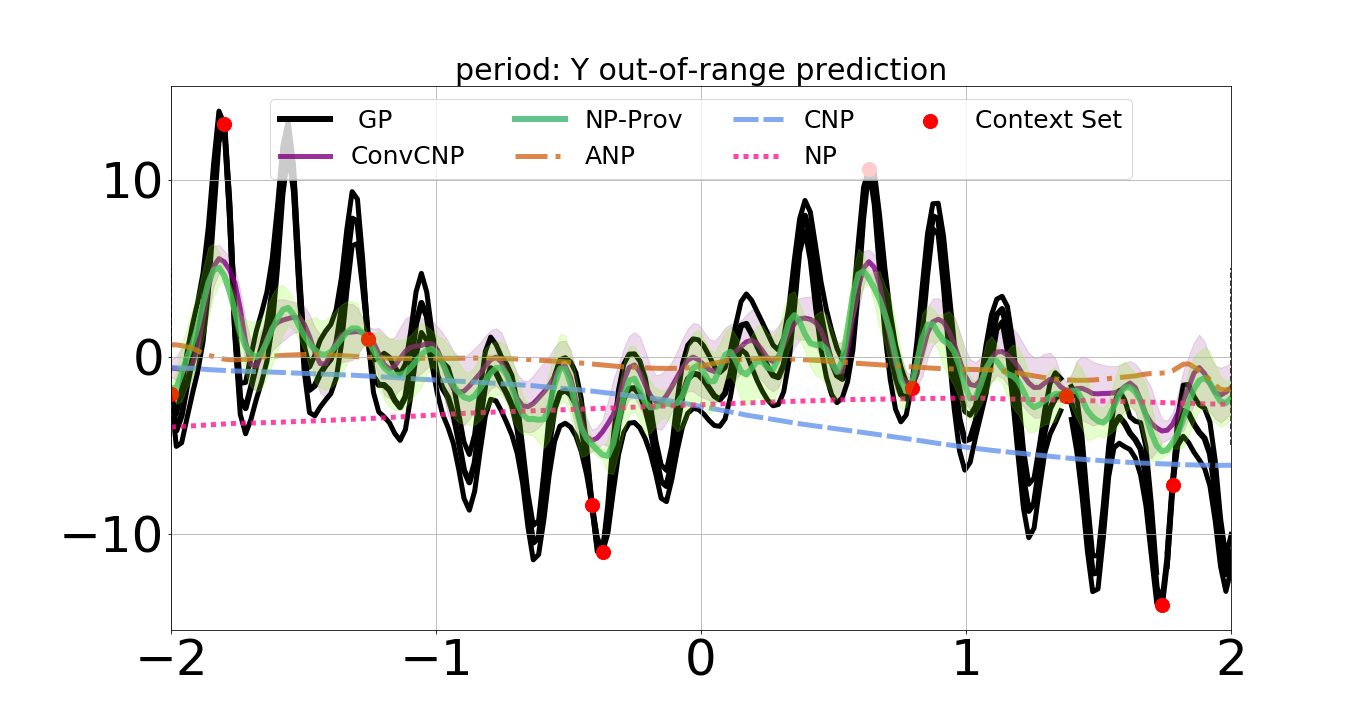}}
        {\includegraphics[width=0.32\linewidth]{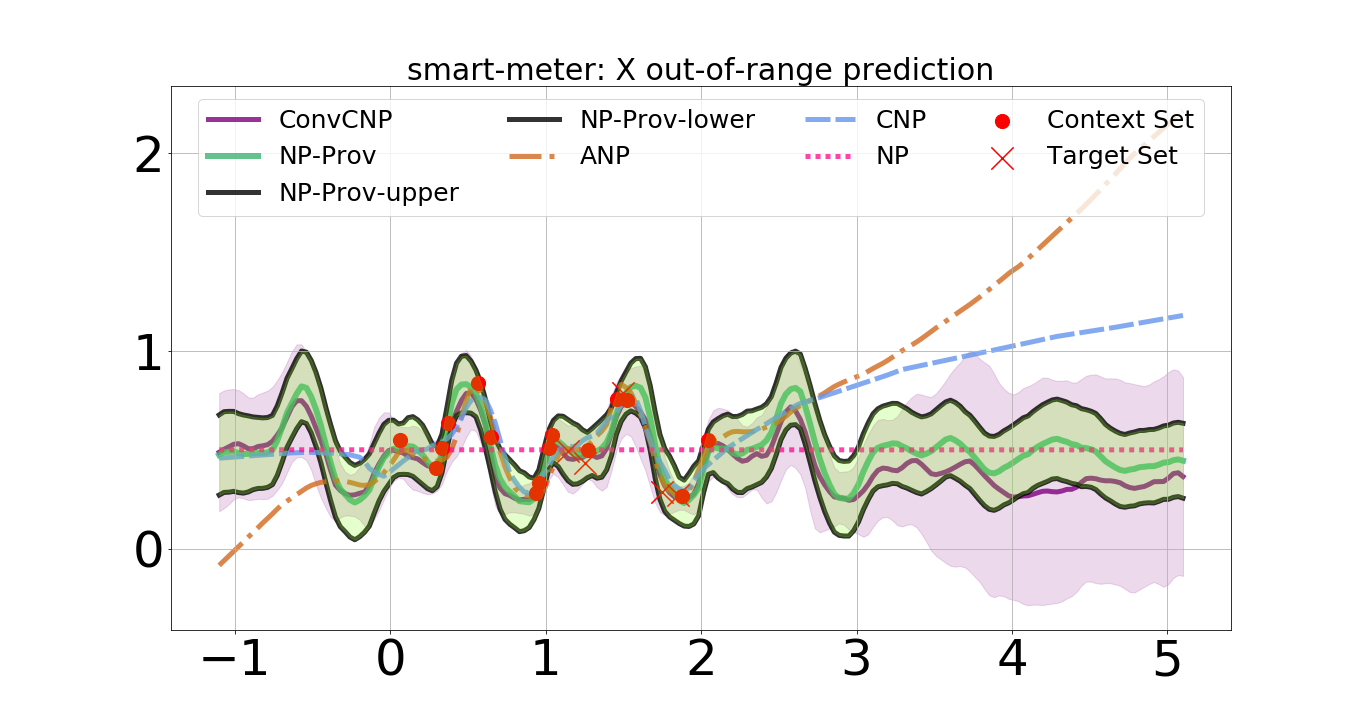}}
        {\includegraphics[width=0.32\linewidth]{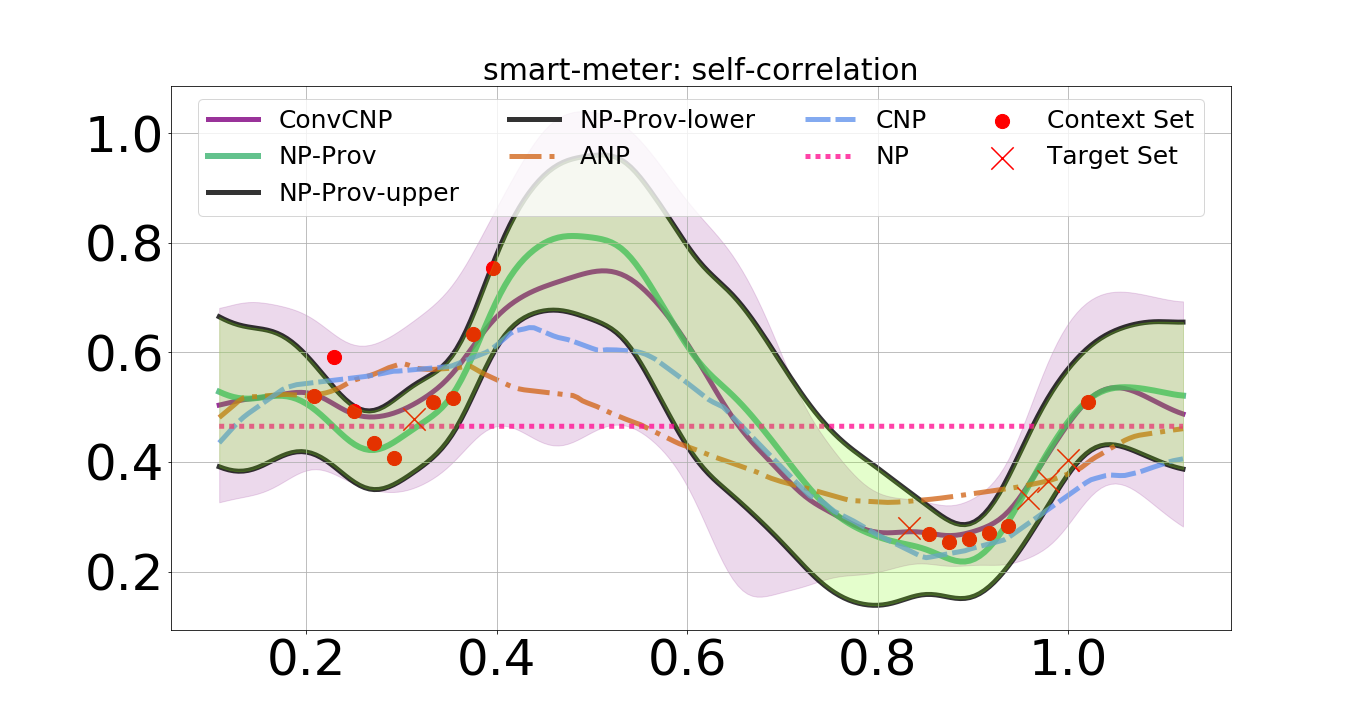}}
        {\includegraphics[width=0.32\linewidth]{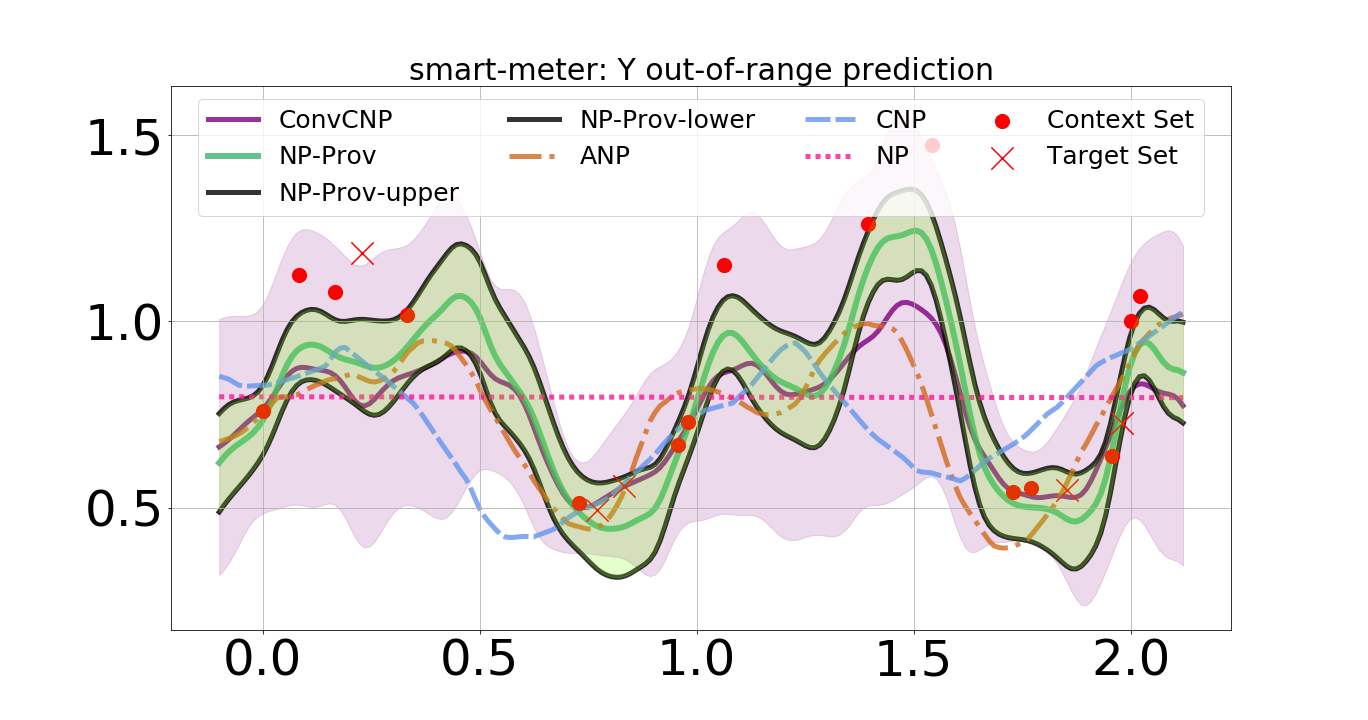}}
    \caption{Model predictions on 4 datasets: EQ (the first row), Matern (the second row), Weakly period (the third row) and Smart meter (the fourth row). The first column is out-of-x-range prediction, the second column is self-correlation prediction, the third column is out-of-y-range prediction.}
    \label{fig: 1d results}
\end{figure}

Table \ref{Table:LL of 1d datasets} shows the log-likelihood (on the probability density function) of five methods. Model performance is presented in Fig \ref{fig: 1d results}, which supplement mean and variance results from the original GP as a reference for synthesized datasets. Table \ref{Table:LL of 1d datasets} shows both ConvCNP and NP-PROV outperform others significantly. The result of NP is quite unstable, due to stochastic encoding of the latent variable. The first column of Fig \ref{fig: 1d results} shows the results with the testing range [-5, 5] for the GP-sampled datasets. NP, CNP, and ANP predict oscillated mean values; therefore, we omit their variance values for display clarity. Both NP-PROV and ConvCNP match well with the original GP. This advantage sources from convolution, where filters can slide along the axis. NP-PROV usually predicts a narrower variance than ConvCNP when the target points are adjacent to context points. This might be attributed to more uncertainty reduction from context self-correlation. When Smart Meter is extended to the interval [-1, 5], the predicted variance becomes much higher than NP-PROV (as it only adopts cross-correlation), and the target points become far from the context points. NP-PROV mitigates this issue by leveraging the self-correlation on target data.

\begin{table}[ht]
  \caption{Log-likelihood of off-the-grid datasets (mean \(\pm\) standard deviation)}
  \label{Table:LL of 1d datasets}
  \centering
  \begin{tabular}{lllll}
    \toprule
    \cmidrule(r){1-5}
    Model &       EQ               & Matern     & Weakly Periodic  &   Smart Meter   \\
    \midrule
    NP  & -1.20 \(\pm\) 0.43 &  -0.82 \(\pm\) 1.95& -1.75 \(\pm\) 1.36 & 0.93 \(\pm\) 0.23\\
    CNP & -1.10 \(\pm\) 0.02 &  -1.36 \(\pm\) 0.03& -2.04 \(\pm\) 0.02 & 1.81 \(\pm\) 0.06\\
    ANP & -0.35 \(\pm\) 0.01 &  -0.75 \(\pm\) 0.02& -2.09 \(\pm\) 0.03 & 1.66 \(\pm\) 0.05\\
    ConvCNP & 2.15 \(\pm\) 0.05 & 0.83 \(\pm\) 0.06& -1.15 \(\pm\) 0.01 & \textbf{2.65 \(\pm\) 0.06}\\
    NP-PROV & \textbf{2.20 \(\pm\) 0.02}  & \textbf{0.90 \(\pm\) 0.03}&  \textbf{-1.00 \(\pm\) 0.02}  &  2.32 \(\pm\) 0.05 \\
    \midrule
    ConvCNP (self) & \textbf{77.60 \(\pm\) 2.33} & 40.26 \(\pm\) 0.81 & -47.63 \(\pm\) 0.75 & 0.95 \(\pm\) 0.00  \\
    NP-PROV (self) & 77.55 \(\pm\) 2.61  & \textbf{44.14 \(\pm\) 1.03} & \textbf{-40.96 \(\pm\) 0.89}  & \textbf{0.99 \(\pm\) 0.01} \\
    \bottomrule
  \end{tabular}
\end{table}

We analyze the impact of self-correlation on the model. In the second column of Fig\ref{fig: 1d results},  we give context data in smaller intervals of [-1, -0.5] \(\cup\) [1.2, 1.7] for GP-sampled datasets and [0.2, 0.4] \(\cup\) [0.8, 1.1] for Smart Meter to get higher self-correlations between context points. Also, in Table \ref{Table:LL of 1d datasets}, we present the log-likelihood of ConvCNP (self) and NP-PROV (self) where target points are adjacent to these compacted regions of context data. NP-PROV predicted lower variances on target points near compacted regions, indicating the ability to capture the self-correlation between context points.

Finally, we modify the testing GP generator to \(y = 10\times\mathcal{GP}(x)\) and 1.5 times of the original Smart Meter to evaluate model performance when testing \textbf{Y} goes beyond the training range. The third column of Fig \ref{fig: 1d results} reveals that all methods are too conservative in adaptation due to the normalization effect of activation layers. However, the variance of NP-PROV on the context data is close to zero, and it is not affected by the explosion of target values, which suffice the hypothesis of a stochastic process. Such an advantage is crucial when there are shifts in time series data.

\subsection{On-the-grid datasets}
Given different proportions of pixel values, the objective is to complete the whole image and estimate uncertainty at unknown pixels. Four image datasets are introduced: MNIST, SVHN, celebA \(32 \times 32\) and miniImageNet \cite{deleu2019torchmeta}. Except MNIST, all the other datasets are RBG images. MiniImageNet is used to verify the model capability of recovering images of unseen classes. The context points are sampled from \(\mathcal{U}(\frac{n_{total}}{100}, \frac{n_{total}}{2})\) and the target points are \(n_{total}\), i.e., the whole image.

\begin{figure}[ht]
    \centering
        \subfigure[]        {\includegraphics[width=0.27\linewidth]{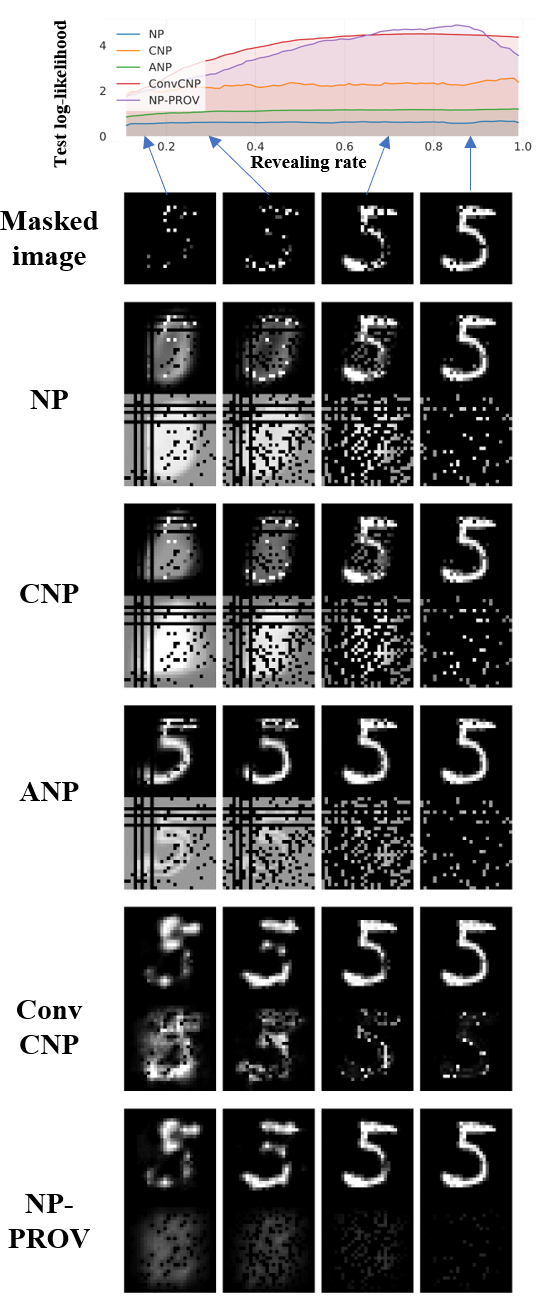}}
        \subfigure[]
        {\includegraphics[width=0.235\linewidth]{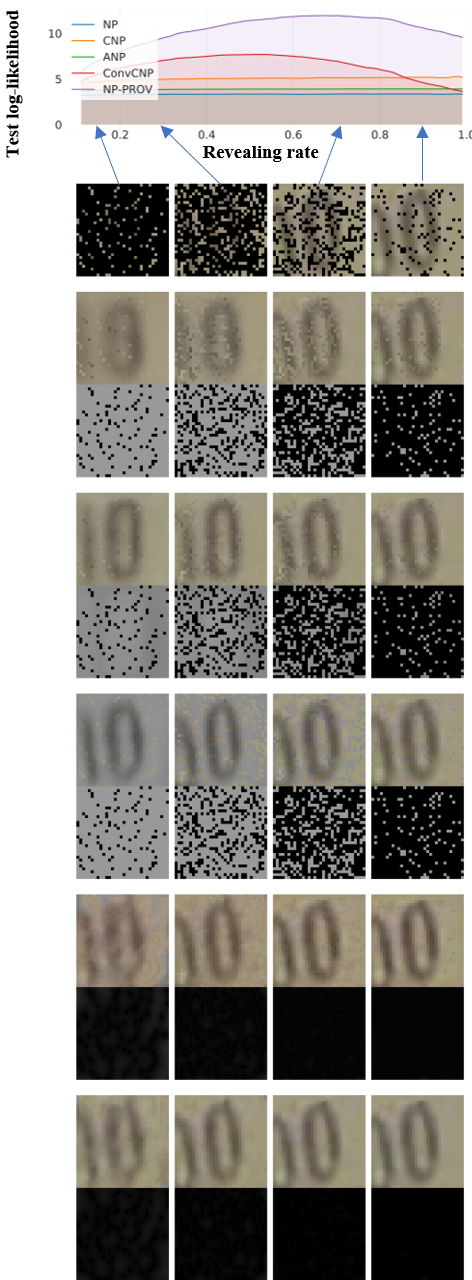}}
        \subfigure[]
        {\includegraphics[width=0.235\linewidth]{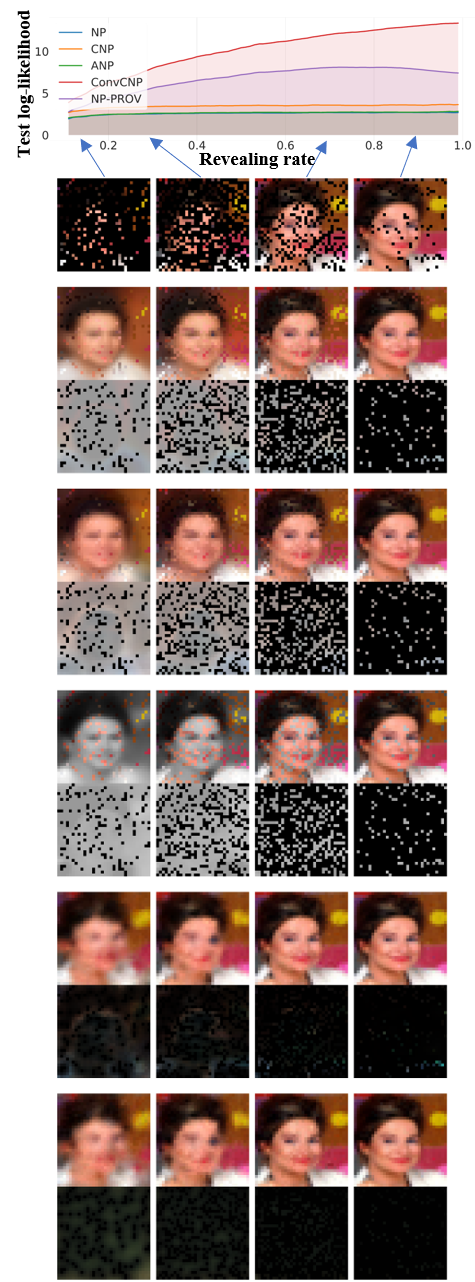}}
        \subfigure[]
        {\includegraphics[width=0.23\linewidth]{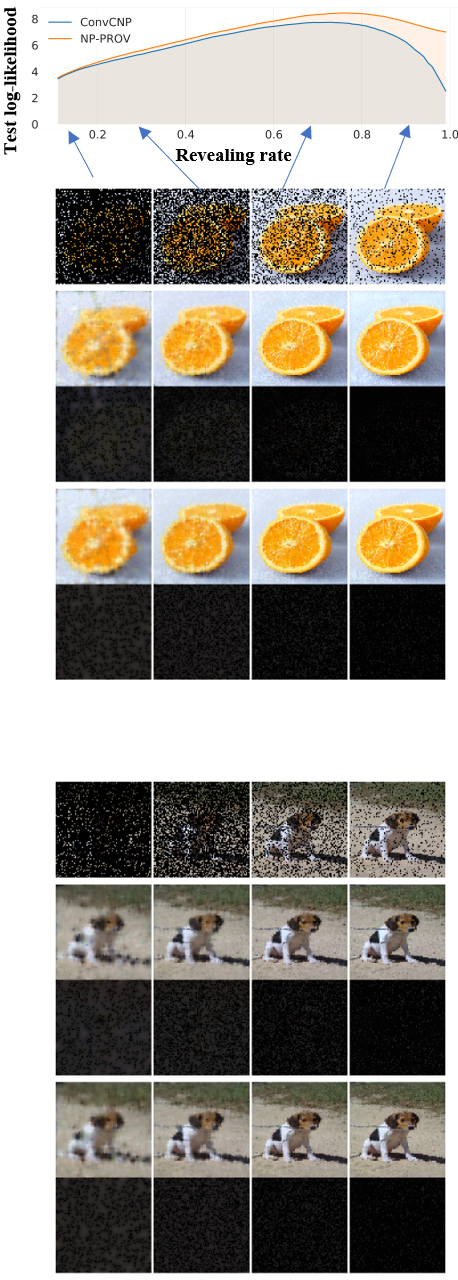}}
    \caption{Test log-likelihood w.r.t revealing rates (the upper row). The lower row presents some samples from (a) MNIST, (b) SVHN, and (c) celebA \(\times\)32 and miniImageNet (d) with different revealing rates: {10\%, 30\%, 70\%, 90\%}. The first row in every method predicts mean and the second row predicts variance.}
    \label{fig: 2d results}
\end{figure}

Table \ref{Table:LL of 2d datasets} demonstrates the log-likelihood of the comparing methods on testing sets. The results of miniImageNet are only displayed for ConvCNP and NP-PROV since other baselines are overwhelmed by the image size \(84 \times 84\). Fig \ref{fig: 2d results} shows the prediction results on the datasets. NP-PROV and miniImageNet massively outperform other baselines. NP-PROV achieves better results on SVHN and miniImageNet. From the variance results in Fig \ref{fig: 2d results}, most of the methods don't exhibit explainable variances in these two methods, because they are similar to "unsupervised" scenarios where one or a few images represent a new class, hence the methods are unable to gain enough variance information from the pixel values. Whereas there are only limited patterns for a digit or an outline of a face in MNIST and celebA, other baselines adopting pixel values can learn a stabilized mean and display variance on the edges.

\begin{table}[ht]
  \caption{Log-likelihood of on-the-grid datasets  (mean \(\pm\) standard deviation)}
  \label{Table:LL of 2d datasets}
  \centering
  \begin{tabular}{lllll}
    \toprule
    \cmidrule(r){1-5}
    Model     & MNIST             & SVHN              & celebA \(32\times 32\)  &   miniImageNet  \\
    \midrule
    NP        & 0.65 \(\pm\) 4e-4  & 3.21 \(\pm\) 6e-4  &  2.78 \(\pm\) 1e-3      &  --      \\
    CNP       & 1.94 \(\pm\) 4e-2  & 4.48 \(\pm\) 5e-3  &  3.09 \(\pm\) 4e-2      &  --      \\
    ANP       & 0.95 \(\pm\) 3e-3  & 3.50 \(\pm\) 5e-3  &  2.32 \(\pm\) 4e-2      &  --      \\
    ConvCNP   & \textbf{2.98 \(\pm\) 4e-2}  & 6.03 \(\pm\) 2e-1  &  \textbf{6.35 \(\pm\) 2e-1}    & 3.65 \(\pm\) 4e-2 \\
    NP-PROV   & 2.66 \(\pm\) 3e-2  & \textbf{8.24 \(\pm\) 5e-2}    & 5.11 \(\pm\) 1e-2   &    
     \textbf{4.39 \(\pm\) 2e-1}\\
    \bottomrule
  \end{tabular}
\end{table}

\section{Related Work}
Since Gaussian processes suffer high computational cost on matrix inverse, recent work focuses on adopting fast forward-feeding neural networks to substitute original GP. A group of Neural Processes-based models is proposed based on variational inference and ELBO (Evidence Lower BOund) \cite{rudner2018connection}. Neural process families abstract a latent variable or a function from context data and decode the function to the target data based on the relationships between target data and context data. The variations of NPs mostly depend on how the relationships are presented. The original NP \cite{garnelo2018neural} and conditional NP\cite{garnelo2018conditional} adopt mean function to extract stochastic and deterministic latent variables that suffice the exchangeability of a stochastic process. Attentive NP \cite{kim2019attentive} considers the self-attention between context points and cross-attention between context points and target points so that attentive aggregation towards latent variables can be used. Regarding nonstructural relationships between data points, Sequential NP \cite{singh2019sequential} and recurrent NP\cite{willi2019recurrent} address the temporal correlations in time series. Convolutional conditional NP \cite{gordon2019convolutional} utilizes the translation equivariance of convolution to predict on out-of-training range target locations. Functional NP \cite{louizos2019functional} and Graph NP \cite{carr2019graph} exploit topological relationships between context and target nodes. Similar to Meta-Agnostic-Meta-Learning(MAML), residual NP\cite{Lee2020ResidualNP} assumes that few shot tasks share a unified latent variable with minor variations on each task. Thus, the latent variable of the context data is adapted based on the prediction residues. Besides, numerous work focuses on enhancing Gaussian Processes with non-linear feature extraction using neural networks. NNGP \cite{lee2017deepneuralnetworks} and ConvNet \cite{garriga2018deepshallow} explore the relationships between Gaussian processes and one layer neural networks from the perspective of Bayesian inference and approximation. Deep Gaussian processes \cite{damianou2013deepgaussianprocess} and deep kernel learning \cite{wilson2016deepkernel} substitute manual designed kernels with neural networks to extract higher-dimensional features.

\section{Conclusions and Discussions}
We introduced NP-PROV, a new member of Neural Processes that derive variances from a position-relevant-only latent space. We verify that NP-PROV can estimate bounded uncertainty when context data has high self-correlations or function values are out-of-the-training range. We mitigate the problem of predicting stabilized variance under shifts in function values. We believe that NP-PROV opens a door of rethinking the relationships between the mean and variance in Neural Processes. This work leaves multiple avenues for future improvements. It would be interesting to see the mean derivation be more adaptive to out-of-the training range. Also, unifying on-and-off-the-grid version of NP-PROV to fit in higher-dimensional space, e.g., by designing a hyper grid for convolution.

\section*{Broader Impact}
The proposed neural processes have implications for both researchers and practitioners who rely on dynamic data or rapidly shifting contexts to make reliable predictions. This work enables a learning model to adapt to new samples in testing environments and, therefore, overcomes the limitations or biases in pre-prepared training data. This work fundamentally supports better modeling that improves all relevant applications that are based on neural processes. It has the potential to mitigate the challenge of model uncertainty posed by the lack of training data, overfitting, and imbalanced training and testing data distribution in traditional machine learning research. Potential applications benefiting from this technology include super-resolution image reconstruction, fine-grained map interpolation in order to save costs from densely distributed sensors, and medical data generation using only a few shots. One may also argue that the related research can be used for military computer vision tasks, high precision GPS search, and drone state estimation under dynamic environments.





\bibliographystyle{apalike}
\bibliography{bio}

\end{document}